\documentclass{article} 
\usepackage{iclr2027_conference,times}

\usepackage{amsmath,amsfonts,bm}

\def\eqref#1{equation~\ref{#1}}

\def\1{\bm{1}}

\DeclareMathAlphabet{\mathsfit}{\encodingdefault}{\sfdefault}{m}{sl}
\SetMathAlphabet{\mathsfit}{bold}{\encodingdefault}{\sfdefault}{bx}{n}

\usepackage{hyperref}
\usepackage{url}
\usepackage{graphicx}
\usepackage{wrapfig}
\graphicspath{{figs/}}
\usepackage{booktabs}
\definecolor{increase}{HTML}{C62828}
\definecolor{decrease}{HTML}{1565C0}
\usepackage[table]{xcolor}
\newcommand{\ours}{DFC}
\usepackage{algorithm}
\usepackage{algpseudocode}
\usepackage{dsfont}
\newcommand{\indic}{\mathds{1}}

\title{Density Feedback Control for Transferability of Generative Sparse Adversarial Attacks}

\author{Dongsu Song \& Jay Hoon Jung \\
Korea Aerospace University \\
Goyang, Republic of Korea
}

\iclrfinalcopy 
\begin{document}

\maketitle
\lhead{Preprint. Under review.} 

\begin{abstract}
Generator-based sparse adversarial attacks train encoder--decoder networks
to produce transferable sparse perturbations for black-box robustness
evaluation. Their perturbation density, however, is set only indirectly by a
fixed sparsity coefficient whose mapping to density is unknown and shifts
during training, so reaching a target density requires re-tuning; this fixed
pressure also yields little transferability gain once density reduction
begins. We propose Density Feedback Control (DFC), which tightens or relaxes
this coefficient at the end of each epoch according to the gap between the
observed hard-mask density and the target, without changing the generator
architecture or loss. At a 10\% target density, DFC brings the mean density
of three base attacks within one percentage point of the target in a single
run and improves the transfer attack success rate (ASR) of two of them on CNN
and Transformer targets, including under per-image $\ell_0$ budgets, while
perturbing fewer pixels. Tracing GPG's training, we observe that as DFC relaxes the coefficient,
the transfer ASR recovers with the density and overtakes the
fixed-coefficient baseline with fewer pixels. For EGS-TSSA, which
restricts perturbations to source-model feature-importance regions, DFC
yields no consistent gain; we analyze how this restriction interacts with
density feedback.
\end{abstract}

\section{introduction}
Sparse adversarial attacks expose a failure mode in which a vision classifier can be deceived by modifying only a small subset of input pixels \citep{sparsefool, greedyfool, tsaa, rfpar, egs-tsaa, segpar, gpg}. Sparse attacks are important for assessing robustness, as they serve as stress tests under threat models that restrict the number of modifiable input elements \citep{sparse-rs}. Sensitivity to perturbations affecting only a few pixels also offers a complementary perspective on the analysis of deep neural networks (DNNs) \citep{greedyfool}.
In particular, generator-based sparse attacks have been proposed to learn a source model’s vulnerabilities using training data and transfer the resulting sparse attacks to a target model, thereby revealing its vulnerabilities in a black-box setting where its internal information is unavailable. 

Such attacks, including TSAA~\citep{tsaa}, EGS-TSSA~\citep{egs-tsaa}, and GPG~\citep{gpg}, train the generator on an adversarial loss plus a sparsity loss weighted by a coefficient $\lambda$, which sets the perturbation density only indirectly. Reaching a desired density therefore requires repeated training runs for TSAA, multi-stage optimization for EGS-TSSA, and fine-tuning with a re-tuned coefficient for GPG. In each case, $\lambda$ is specified before a training or fine-tuning stage, whereas the resulting density is observed only afterwards. The mapping from $\lambda$ to the density is unknown, differs across attacks, and evolves during generator training, preventing this open-loop procedure from directly targeting a prescribed density.

\begin{figure}[t]
    \centering
    \includegraphics[width=\linewidth]
    {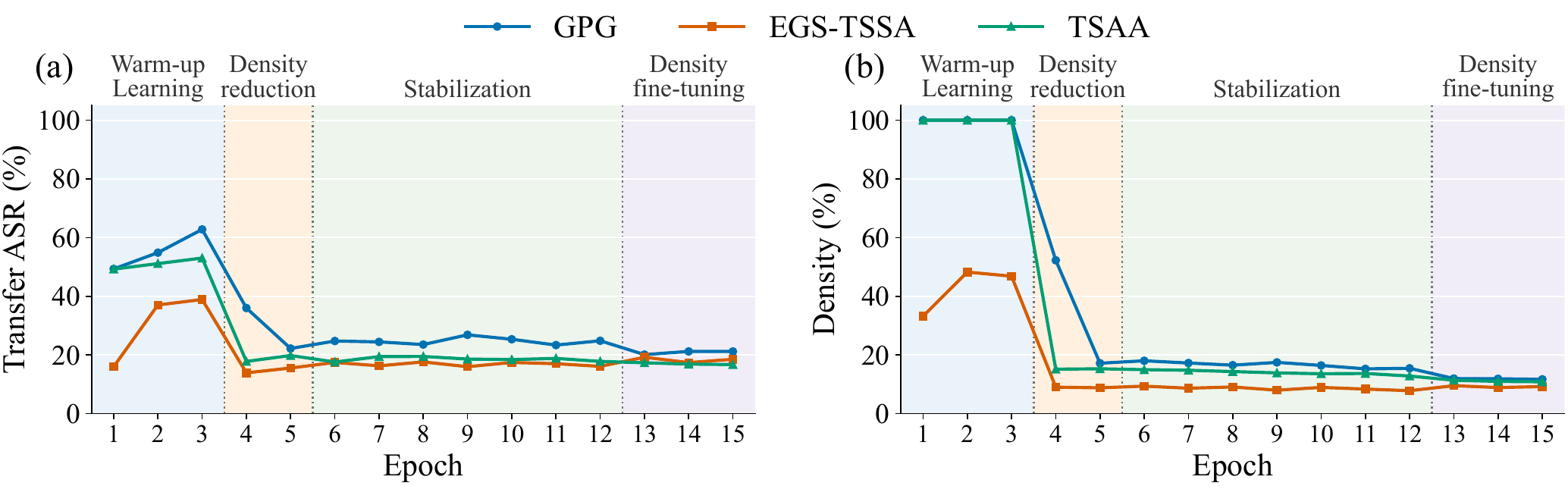}
    \vspace{-18pt}
    \caption{
        Training dynamics of generator-based sparse adversarial attacks.
        Evolution of (a) transferability and (b) mean pixel density
        across four training stages: warm-up, density reduction,
        stabilization, and density fine-tuning.
        Transferability is measured by the mean attack success rate
        (ASR) across five CNN-based and five Transformer-based
        target models.
    }
    \vspace{-12pt}
    \label{fig:training_dynamics}
\end{figure}

Beyond making the density difficult to prescribe, fixed-coefficient training also
limits transferability. Fig.~\ref{fig:training_dynamics} traces GPG, TSAA, and
EGS-TSSA under the protocol of \citet{gpg}, which proceeds through warm-up, density
reduction, stabilization, and density fine-tuning with a re-tuned coefficient.
Transferability improves primarily during warm-up and, as expected, drops during
density reduction; however, the stabilization and fine-tuning stages recover little
of this loss despite occupying more than half of training, while the density remains nearly constant
(Fig.~\ref{fig:training_dynamics}b). Because the coefficient is fixed within each stage and changed only by manual
re-tuning between stages, the sparsity pressure does not respond to where the
density lies relative to the target. The generator nonetheless keeps changing
under this pressure: it replaces over 30\% of its support per epoch
and discards locations more effective than random ones, a loss its ASR
never recovers (Sec.~\ref{sec:dynamics}). This motivates making the sparsity
pressure respond to the observed density, tightening it when the density is
above the target and relaxing it when it falls below.

We propose Density Feedback Control (\ours{}), which links the weight of the
differentiable sparsity loss to the hard-mask density that is actually
evaluated. At the end of each epoch, \ours{} measures the mean hard-mask density
of the generated perturbations and updates the sparsity coefficient from its
normalized deviation from the target. \ours{} changes only how the coefficient
is set; the architecture and loss of the base attack are unchanged, so it
replaces the multi-stage schedules of existing generator-based attacks without
further modification.

Our contributions are summarized as follows:
\begin{itemize}
  \item \ours{} treats the sparsity coefficient as a control input driven by the
hard-mask density, bringing the mean density of three base attacks to within
one percentage point of a prescribed target in a single training run, without
stage-wise re-tuning.
\item \ours{} improves the mean transfer ASR of GPG and TSAA on CNN and
Transformer targets while perturbing fewer pixels on average. Under per-image $\ell_0$ budgets, its gain comes not only from placing more
images within the budget but also from attacking the same images more
successfully.
\item Tracing training, we observe that as \ours{} relaxes the coefficient,
the density recovers from below and the transfer ASR recovers with it,
overtaking the fixed-coefficient baseline with fewer pixels.
\end{itemize}

\section{Related works}

Black-box adversarial attacks, which do not require access to
the target model's internal parameters or gradients, can be
broadly categorized into query-based and transfer-based methods.

\subsection{Query-based Sparse Attacks}
Query-based attacks craft adversarial examples by iteratively modifying the input,
guided by the target model's prediction scores or labels. Early sparse variants used
evolutionary search over individual pixels \citep{onepixel} or over localized
perturbations parameterized as lines and curves \citep{scratchthat}. Later work
searched for vulnerable patches with reinforcement learning \citep{patchattack},
rearranged a small number of pixels \citep{pixle}, and selected pixel replacements
from a synthetic image via Bayesian search under an $\ell_0$ constraint
\citep{brusleattack}. \citet{rfpar} introduced a REINFORCE-based framework with
remember-and-forget processes for classification and object detection, which
\citet{segpar} extended to decision-based attacks on semantic segmentation.
Because these methods require repeated queries to the target model, transfer-based
approaches instead craft perturbations on a source model without target-model
feedback.

\subsection{transfer-based sparse attacks}
\label{sec:transfer}
Transfer-based sparse attacks generate adversarial perturbations against a source model under white-box access and transfer them to a target model whose internal information is unavailable. PGD$_{0}$ adapted projected gradient descent (PGD) to sparse attacks by projecting perturbations onto an $\ell_0$ ball \citep{pgd, pgd0}. SparseFool extended DeepFool, which seeks perturbations that cross a nearby decision boundary, to the sparse setting \citep{sparsefool, deepfool}. GreedyFool uses input gradients to greedily select influential pixels and subsequently prunes redundant perturbations to reduce the number of modified pixels \citep{greedyfool}. More recently, sPGD jointly optimized perturbation magnitudes and a sparse support mask through PGD-like gradient updates under an $\ell_0$ constraint \citep{sPGD}. Despite their strong attack performance on the source model, these gradient-based methods exhibit limited transferability to unseen target models.

Generator-based adversarial attacks train a generator to produce adversarial perturbations with improved transferability \citep{cross, advgan, ATN, gap, CAE}. Inspired by this paradigm, \citet{tsaa} proposed TSAA, which encodes the input image into a compact representation and learns a pixel-selection mask $M$ and perturbation values $P$ to construct a sparse perturbation $\delta$. TSAA enables substantially faster attack generation than gradient-based methods while improving transferability. Building on this approach, EGS-TSSA \citep{egs-tsaa} improves transferability by restricting the selection of $M$ to regions containing features important for classification, using a Group Feature Importance (GFI) mask derived from the feature maps of the source model. Similarly, GPG \citep{gpg} enhances transferability by encouraging $P$ to resemble class-specific perturbations generated by PGD.

The training objective for a sparse perturbation generator is:
\begin{equation}
\mathcal L(\theta;\lambda)
=
\mathcal L_{\mathrm{adv}}(\theta)
+
\lambda\mathcal L_{\mathrm{spa}}(\theta)
,
\end{equation}
where $\mathcal{L}_{\mathrm{adv}}$ encourages misclassification, $\mathcal{L}_{\mathrm{spa}}$ promotes sparsity, and $\lambda$ controls their relative importance. Prior methods have primarily improved transferability through the design of
$\mathcal{L}_{\mathrm{adv}}$.

\subsection{Adaptive Control of Regularization Weights}

The weight of a regularization term can be adjusted during training by a prescribed
schedule or by feedback. Prescribed schedules, such as cyclical KL annealing~\citep{Fu}, repeatedly vary the regularization weight during training.
Feedback methods treat the weight as a Lagrange multiplier updated from constraint
violation~\citep{MDMM, GECO}, with proportional--integral~\citep{ControlVAE} or
PID~\citep{PIDLagrangian} variants that suppress oscillations and converge smoothly to
the target.

Target sparsity has also been pursued through adaptive sparsity losses~\citep{Retsinas} and constrained optimization \citep{GallegoPosada}. In sparse adversarial
attacks, per-image methods such as PGD$_0$ enforce the $\ell_0$ budget directly by
projection, and \citet{Homotopy} replace a fixed sparsity weight with a decreasing
sequence of weights. For generator-based attacks, whose density is a property of the
learned mapping, $\lambda$ is still set manually, as reviewed in
Sec. ~\ref{sec:transfer}; to our knowledge, none controls the density by feedback
during training.

\ours{} brings such feedback to generator-based sparse attacks.
As in proxy-Lagrangian methods \citep{cotter} and unlike loss-level
controllers \citep{ControlVAE}, the
generator is trained on a differentiable surrogate while the coefficient responds to
the observed hard-mask density, the quantity that is actually evaluated.

\begin{figure}[t]
    \centering
    \includegraphics[width=\linewidth]{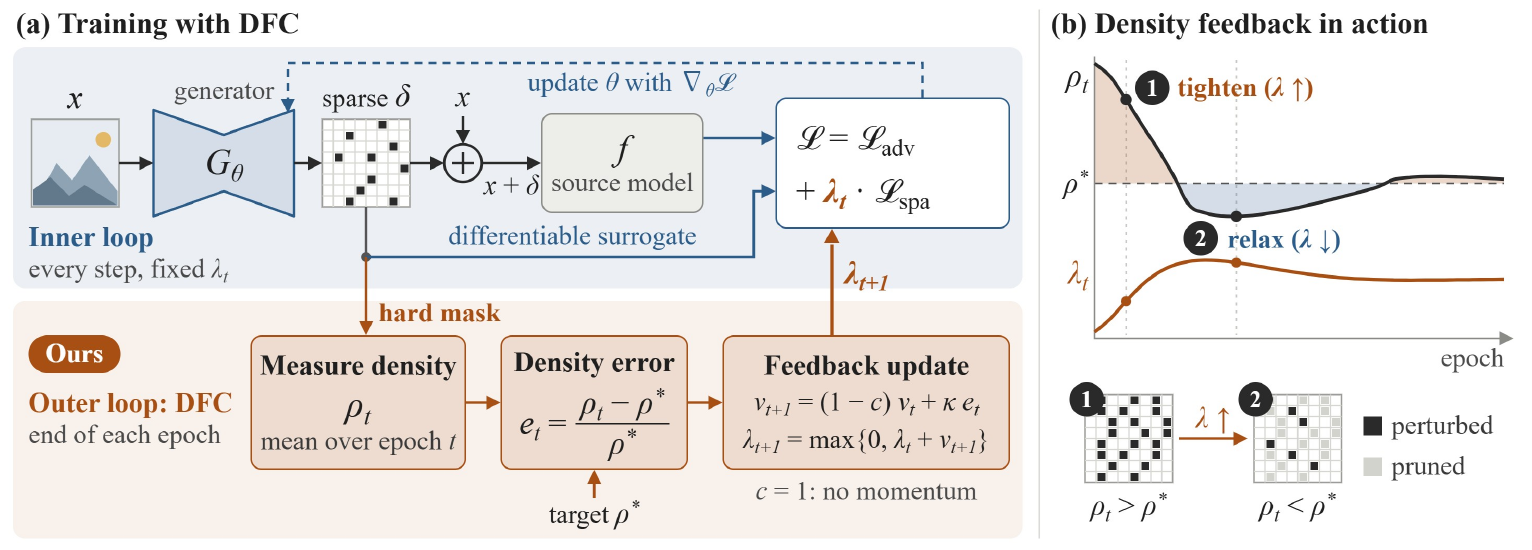}
    \vspace{-.8cm}
    \caption{Overview of \ours{}. (a) The generator $G_\theta$ of the base attack is
    trained at every step on $\mathcal{L}_{\mathrm{adv}} + \lambda_t
    \mathcal{L}_{\mathrm{spa}}$, with $\lambda_t$ fixed within epoch $t$. At the end of each epoch, \ours{} measures
    the mean hard-mask density $\rho_t$, computes its normalized error from the target
    $\rho^*$, and updates the sparsity coefficient. (b) Schematic training dynamics:
    the sparsity pressure tightens while $\rho_t > \rho^*$ and relaxes once $\rho_t$
    falls below $\rho^*$, so the mean density approaches the target. Masks illustrate
    one image; \ours{} regulates the mean density.}
    \label{fig:overview}
    \vspace{-.4cm}
\end{figure}

\section{Method}
\subsection{Problem Formulation}
Given a target mean pixel density $\rho^*$, we seek a generator that maximizes attack
effectiveness while producing perturbations at that density:
\begin{equation}
\min_{\theta}\ \mathcal{L}_{\mathrm{adv}}(\theta)
\quad \text{s.t.} \quad \rho(\theta)=\rho^*,
\qquad
\rho(\theta)=\mathbb{E}_{x}\Big[\tfrac{1}{HW}\sum_{i,j}\indic\big[\|\delta_{:,i,j}(x)\|_0>0\big]\Big].
\end{equation}
Because $\rho(\theta)$ is computed from the hard mask and is non-differentiable, existing
methods instead minimize the surrogate $\mathcal{L}_{\mathrm{adv}}+\lambda\mathcal{L}_{\mathrm{spa}}$
with a fixed $\lambda$. The map $\lambda\mapsto\rho$ is unknown, attack-specific, and
non-stationary during training, so a fixed $\lambda$ leaves $\rho$ uncontrolled.
We instead treat $\lambda$ as a control input and adjust it from the observed density
error, closing the loop between the differentiable surrogate and the density that is
actually evaluated.

\subsection{Density Feedback Control}
\label{sec:dfc}

\paragraph{Density measurement.}
\ours{} updates the sparsity coefficient once per epoch from the observed density.
Let $\rho_t$ denote the mean density during epoch $t$, i.e., the average fraction of
active spatial locations in the hard masks of all training samples processed in that
epoch, where a location counts as active if any color channel is perturbed. Since the
masks are already computed in the forward pass, $\rho_t$ is obtained at no additional
cost. Given a target density $\rho^*>0$, the normalized density error is
\begin{equation}
e_t=\frac{\rho_t-\rho^*}{\rho^*}.
\end{equation}

\paragraph{Controller.} The generator is trained with a fixed $\lambda_t$ throughout
epoch $t$. At the end of the epoch, DFC updates a velocity state $v_t$ and then the
coefficient:
\begin{equation}
v_{t+1}=(1-c)\,v_t+\kappa e_t,\qquad \lambda_{t+1}=\max\{0,\lambda_t+v_{t+1}\},
\label{eq:dfc}
\end{equation}
where $\kappa > 0$ is the feedback gain and $c \in (0, 1]$ is the damping coefficient.
A density above (below) the target pushes the coefficient up (down). While the clamp is inactive, eliminating $v_t$ gives
$\lambda_{t+1} = \lambda_t + \kappa e_t + (1-c)(\lambda_t - \lambda_{t-1})$: the
coefficient follows the density error as in the basic multiplier update of
\citet{MDMM}, plus a momentum term that carries the previous adjustment
forward \citep{polyak}. Setting $c = 1$ removes the momentum and
recovers the basic update. Unlike dual ascent on a single Lagrangian, the generator
minimizes the differentiable surrogate $L_{\mathrm{spa}}$ while the coefficient
responds to the hard-mask density, as in proxy-Lagrangian methods
\citep{cotter}. Because $\rho_t \in [0, 1]$, the error is bounded below by $-1$
but can be large above the target (up to 9 for $\rho^* = 10\%$); the error term
therefore tightens the pressure quickly when the density is too high and relaxes it
gradually when it is too low.

\paragraph{Gain and initialization.}
Since $e_t$ is dimensionless, $\kappa$ sets the scale of each adjustment. We match it to
the order of magnitude of the initial coefficient,
\begin{equation}
\kappa=10^{\lfloor\log_{10}\lambda_0\rfloor},
\end{equation}
so that the adjustments are commensurate with the coefficient range of each base
attack. This allows the same $c$ to be shared across base attacks whose coefficients differ in scale, leaving $c$ and $\lambda_0$ as the only hyperparameters
of \ours{}. We initialize $v_0=0$. The controller runs outside the generator's
computational graph (see Fig. \ref{fig:overview} and Algorithm~\ref{alg:dfc}).

\section{Experiments}
\subsection{Experimental setting}
\paragraph{Setups.}
We conducted our experiments on the widely used ImageNet dataset \citep{imagenet}. Following \citet{gpg}, we used ResNet-50 as the source model and DenseNet-161, VGG-16, Inception-v3, Wide ResNet-50, EfficientNet-B6, ViT-Small, DeiT-Small, TNT-Small, Twins-Small, and Swin-Tiny as the target models \citep{densenet, vgg, inception, wideresnet, effnet, vit, deit, tnt, twins, swin}. We set $\epsilon=10/255$.

\paragraph{Benchmark Dataset.}
Generator-based sparse attacks, including DFC, were trained on the ImageNet training set and evaluated on the entire ImageNet validation set. All other methods were evaluated on a subset of 5,000 images randomly sampled from the ImageNet validation set using a fixed random seed of 0.

\paragraph{Evaluation Metrics and Comparative Methods.}
Sparsity is defined as the average pixel density of adversarial perturbations. To evaluate transferability, we use the attack success rate (ASR), measured on images correctly classified by each victim model. To assess the effectiveness of DFC, we primarily compare TSAA, EGS-TSSA, and GPG with their respective DFC-enhanced variants. For a more comprehensive evaluation, we additionally include PGD$_0$, sPGD, SparseFool, and GreedyFool as baselines. For these methods, we use the official implementations with default hyperparameter settings, except for those controlling perturbation density.

\paragraph{Generator details.}
We follow the architectures and loss functions of the official implementations
and the training procedure of \citet{gpg}. All generator-based methods, including their DFC variants, are trained for 15 epochs, starting with a warm-up with $\lambda = 0$ of three epochs for the
baselines and two for the DFC variants. The baselines then use fixed
coefficients for reduction and stabilization (nine epochs) and fine-tuning
(three epochs) toward a target density of 10\%, the density at which
\citet{gpg} compare sparse attacks. For the DFC variants, DFC replaces these
phases with feedback toward the same target, $\rho^* = 10\%$; for TSAA, we lower
$\lambda_0$ because the default drove the density to zero after warm-up.
Coefficients and DFC hyperparameters are given in App.~\ref{app:gen}.

\begin{table*}[h]
\centering
\caption{Transfer ASR (\%) on CNN models and mean density (\%).
$^{*}$ denotes source-model results.
Mean ASR is averaged over the five CNN target models, excluding the source model.
Bold indicates the higher ASR within each baseline--DFC pair.}
\label{tab:baseline-dsc-cnn}
\setlength{\tabcolsep}{4pt}
\renewcommand{\arraystretch}{1.12}
\resizebox{\textwidth}{!}{%
\begin{tabular}{lrrrrrrrr}
\toprule
Method & Density & ResNet-50 & IncV3 & VGG16 & Dense161 & WideRes50 & EffNetB6 & Mean ASR \\
\midrule
PGD$_0$ ($100\times5$)
& 10.00 & 100.00$^{*}$ & 10.09 & 20.94 & 23.54 & 29.27 & 10.09 & 18.78 \\
sPGD$_{proj}$ ($T=500$)
& 9.99 & 99.95$^{*}$ & 6.93 & 16.25 & 18.35 & 22.59 & 8.16 & 14.46 \\
sPGD$_{unproj}$ ($T=500$)
& 9.99 & 99.95$^{*}$ & 9.93 & 21.98 & 26.08 & 32.99 & 10.39 & 20.27 \\
SparseFool ($\lambda=20$)
& 12.90 & 100.00$^{*}$ & 9.04 & 15.83 & 15.53 & 18.24 & 9.74 & 13.67 \\
GreedyFool ($\kappa=30$)
& 12.72 & 100.00$^{*}$ & 9.01 & 14.95 & 19.83 & 24.40 & 9.80 & 15.60 \\
\midrule
GPG
& 11.69 & 80.67$^{*}$ & \textbf{18.44} & \textbf{42.49} & 35.80 & 54.73 & 13.50 & 32.99 \\
\rowcolor{gray!15}
GPG-DFC ($c=0.5$)
& 10.82 & \textbf{83.12}$^{*}$ & 13.52 & 37.89 & \textbf{62.90} & \textbf{73.75} & \textbf{22.41} & \textbf{42.09} {\small\textcolor{increase}{(+27.58\%)}} \\
\midrule
TSAA
& 10.86 & 75.41$^{*}$ & \textbf{13.47} & \textbf{39.45} & 28.71 & 32.30 & 12.22 & 25.23 \\
\rowcolor{gray!15}
TSAA-DFC ($c=0.5$)
& 9.99 & \textbf{84.63}$^{*}$ & 11.92 & 38.60 & \textbf{59.13} & \textbf{75.33} & \textbf{19.15} & \textbf{40.83} {\small\textcolor{increase}{(+61.81\%)}} \\
\midrule
EGS-TSSA
& 10.22 & 76.81$^{*}$ & \textbf{17.75} & 44.63 & 32.16 & 35.36 & \textbf{16.32} & 29.24 \\
\rowcolor{gray!15}
EGS-TSSA-DFC ($c=0.5$)
& 10.89 & \textbf{87.54}$^{*}$ & 14.99 & \textbf{45.13} & \textbf{34.53} & \textbf{39.41} & 13.60 & \textbf{29.54} {\small\textcolor{increase}{(+1.00\%)}} \\
\bottomrule
\end{tabular}%
}
\vspace{-.4cm}
\end{table*}

\subsection{Transferability on CNN models}
Table~\ref{tab:baseline-dsc-cnn} reports the transfer ASR of sparse adversarial
examples crafted on ResNet-50 and evaluated on five unseen CNN models.
Conventional sparse attacks (PGD$_0$, sPGD, SparseFool, and GreedyFool) achieve
near-perfect white-box ASR but at most 20.27\% mean transfer ASR, indicating
that their perturbations overfit to the source model. Transfer-oriented methods
(GPG, TSAA, and EGS-TSSA) alleviate this issue (25.23--32.99\%) at comparable
sparsity. DFC substantially improves both GPG (32.99\% $\rightarrow$ 42.09\%)
and TSAA (25.23\% $\rightarrow$ 40.83\%) while lowering their mean density by
0.87~pp, so the gains do not come from perturbing more pixels. The gains are
largest on Dense161 and WideRes50, whereas DFC lowers the ASR on IncV3 and VGG16
(by up to 4.92~pp). In contrast, DFC barely helps EGS-TSSA (29.24\%
$\rightarrow$ 29.54\%), which extends TSAA with a GFI mask for transferability;
as a result, the plain TSAA-DFC outperforms EGS-TSSA with or without DFC. This
suggests that the GFI mask and DFC are not complementary, which we analyze in
Sec.~\ref{sec:gfi}. 

\subsection{Transferability on Transformer models}

Table~\ref{tab:baseline-dsc-transformer} reports the transfer ASR from ResNet-50
to five Transformer models. Transferring across architecture families is
considerably more challenging: conventional sparse attacks reach at most 6.93\%
mean ASR, and transfer-oriented methods at most 9.08\%. In this setting, DFC
yields even larger relative gains than on CNNs, raising the mean ASR of GPG from
9.08\% to 13.45\% (+48.01\%) and that of TSAA from 7.48\% to 12.65\%
(+69.25\%). The gains are largest on ViT-S, DeiT-S, and TNT-S; on TNT-S, DFC
more than doubles the ASR of both attacks. In contrast, DFC lowers the ASR on
Swin-T for both attacks and on Twins-S for GPG (by up to 1.94~pp). For
EGS-TSSA, DFC slightly reduces the mean ASR (Sec.~\ref{sec:gfi}).
\textbf{Overall, DFC alone narrows the gap between TSAA and GPG, proposed
4 years later, from 7.76 to 1.26~pp on CNN models and from 1.60 to 0.80~pp
on Transformer models.}
\begin{table*}[h]
\vspace{-12pt}
\centering
\caption{Transfer ASR (\%) on Transformer models and mean density (\%).
$^{*}$ denotes source-model results.
Mean ASR is averaged over the five Transformer target models, excluding the source model.
Bold indicates the higher ASR within each baseline--DFC pair.}
\label{tab:baseline-dsc-transformer}
\setlength{\tabcolsep}{4pt}
\renewcommand{\arraystretch}{1.12}
\resizebox{\textwidth}{!}{%
\begin{tabular}{lrrrrrrrr}
\toprule
Method & Density & ResNet-50 & ViT-S & DeiT-S & TNT-S & Swin-T & Twins-S & Mean ASR \\
\midrule
PGD$_0$ ($100\times5$) & 10.00 & 100.00$^{*}$ & 6.73 & 4.37 & 7.97 & 7.34 & 7.83 & 6.85 \\
sPGD$_{proj}$ ($T=500$) & 9.99 & 99.95$^{*}$ & 5.69 & 3.67 & 6.45 & 6.45 & 5.99 & 5.65 \\
sPGD$_{unproj}$ ($T=500$) & 9.99 & 99.95$^{*}$ & 6.49 & 4.69 & 7.76 & 7.57 & 8.13 & 6.93 \\
SparseFool ($\lambda=20$) & 12.90 & 100.00$^{*}$ & 5.77 & 2.72 & 5.54 & 4.07 & 4.82 & 4.58 \\
GreedyFool ($\kappa=30$) & 12.72 & 100.00$^{*}$ & 6.73 & 3.74 & 6.19 & 6.20 & 6.69 & 5.91 \\
\midrule
GPG & 11.69 & 80.67$^{*}$ & 8.78 & 5.60 & 10.49 & \textbf{9.43} & \textbf{11.12} & 9.08 \\
\rowcolor{gray!15}
GPG-DFC ($c=0.5$) & 10.82 & \textbf{83.12}$^{*}$ & \textbf{16.81} & \textbf{9.64} & \textbf{23.58} & 7.49 & 9.70 & \textbf{13.45} {\small\textcolor{increase}{($+48.01\%$)}} \\
\midrule
TSAA & 10.86 & 75.41$^{*}$ & 9.77 & 4.68 & 7.21 & \textbf{8.06} & 7.66 & 7.48 \\
\rowcolor{gray!15}
TSAA-DFC ($c=0.5$) & 9.99 & \textbf{84.63}$^{*}$ & \textbf{16.22} & \textbf{9.95} & \textbf{19.20} & 7.48 & \textbf{10.41} & \textbf{12.65} {\small\textcolor{increase}{($+69.25\%$)}} \\
\midrule
EGS-TSSA & 10.22 & 76.81$^{*}$ & \textbf{10.07} & \textbf{4.52} & \textbf{10.64} & \textbf{9.54} & \textbf{9.58} & \textbf{8.87} \\
\rowcolor{gray!15}
EGS-TSSA-DFC ($c=0.5$) & 10.89 & \textbf{87.54}$^{*}$ & 9.52 & 4.52 & 10.63 & 8.45 & 9.23 & 8.47 {\small\textcolor{decrease}{($-4.54\%$)}} \\
\bottomrule
\end{tabular}%
}
\vspace{-12pt}
\end{table*}

\begin{figure}[t]
\centering
\includegraphics[width=\linewidth]{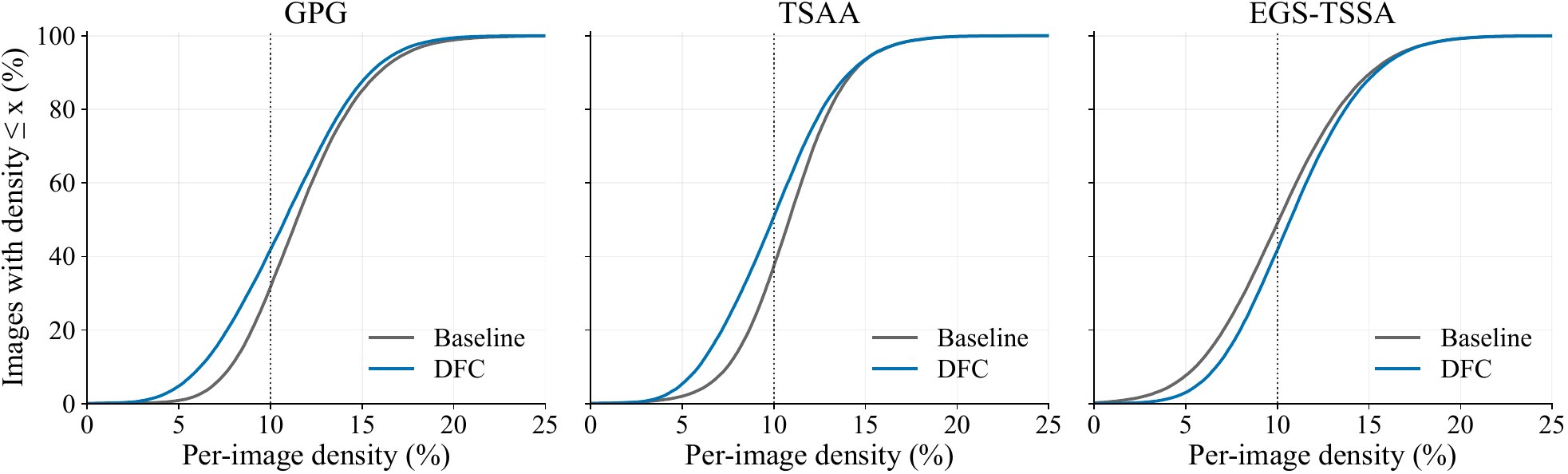}
\vspace{-18pt}
\caption{Per-image perturbation density over all 50{,}000 validation images,
without success filtering. The dotted line marks the 10\% budget.}
\vspace{-12pt}
\label{fig:cdf}
\end{figure}

\subsection{Transferability under Per-Image $\ell_0$ Budgets}
\label{sec:per_image}

Prior generator-based sparse attacks measure sparsity as the average density over
the dataset, which allows more pixels on difficult images as long as the average
stays within the budget. In contrast, query-based sparse attacks impose a strict
$\ell_0$ budget on each image, which better reflects practical threat
models~\citep{rfpar}. Following this protocol, Table~\ref{tab:sparsity_transfer} reports the transfer
ASR averaged over the ten victim models, counting an attack as successful only if
its perturbation density is within the budget; gradient-based attacks are run
separately for each budget. Under this protocol, the fixed-coefficient generator baselines, which lead the
gradient-based attacks in Tables~\ref{tab:baseline-dsc-cnn}
and~\ref{tab:baseline-dsc-transformer}, fall behind them: at $\ell_0 \le 10\%$,
GPG and TSAA achieve 7.95\% and 6.41\%, whereas PGD$_0$ and
sPGD$_{\text{unproj}}$ reach 12.82\% and 13.60\%.
\begin{wraptable}{r}{0.55\textwidth}
\vspace{-0.6cm}
  \centering
  \caption{In-budget transfer ASR (\%) under per-image $\ell_0$ budgets.}
  \label{tab:sparsity_transfer}
  \setlength{\tabcolsep}{2pt}
  \renewcommand{\arraystretch}{1.1}
  \resizebox{\linewidth}{!}{
 \begin{tabular}{lccc}
\toprule
Method & $\ell_{0} \leq 10\%$ & $\ell_{0} \leq 11\%$ & $\ell_{0} \leq 12\%$ \\
\midrule
PGD$_{0}$ & 12.82 & 13.12 & 13.65 \\
sPGD$_{proj}$ & 10.06 & 10.36 & 10.75 \\
sPGD$_{unproj}$ & 13.60 & 14.25 & 14.81 \\
\midrule
GPG & 7.95 & 10.81 & 13.51 \\
\rowcolor{gray!7.5}
GPG-DFC & 13.74\,{\small\textcolor{increase}{(+72.68\%)}} & 16.82\,{\small\textcolor{increase}{(+55.61\%)}} & 19.63\,{\small\textcolor{increase}{(+45.30\%)}} \\
\midrule
TSAA & 6.41 & 8.94 & 11.41 \\
\rowcolor{gray!7.5}
TSAA-DFC & 15.72\,{\small\textcolor{increase}{(+145.31\%)}} & 19.25\,{\small\textcolor{increase}{(+115.34\%)}} & 22.19\,{\small\textcolor{increase}{(+94.55\%)}} \\
\midrule
EGS-TSSA & 9.48 & 11.57 & 13.43 \\
\rowcolor{gray!7.5}
EGS-TSSA-DFC & 8.96\,{\small\textcolor{decrease}{(-5.54\%)}} & 11.16\,{\small\textcolor{decrease}{(-3.49\%)}} & 13.30\,{\small\textcolor{decrease}{(-0.97\%)}} \\
\bottomrule
\end{tabular}
  }
  \vspace{-0.4cm}
\end{wraptable}
Generator-based attacks, including their DFC variants, control density only on
average (Fig.~\ref{fig:cdf}); for the baselines, over 60\% of GPG and
TSAA perturbations exceed the 10\% budget and thus count as failures. DFC puts
GPG and TSAA back in the lead: their DFC variants outperform all baselines under
every budget, improving their in-budget ASR by 5.79--6.12 and 9.31--10.78~pp,
respectively. Notably, TSAA-DFC under the 10\% budget already surpasses every
baseline under the 12\% budget. In contrast, DFC does not improve EGS-TSSA.

\begin{wraptable}{r}{0.55\textwidth}
\vspace{-0.9cm}
  \centering
  \caption{Decomposition of the in-budget ASR gain of DFC at $\ell_0 \le 10\%$ (pp),
averaged over the ten victims. $C_{\mathrm{in}}$ ($C_{\mathrm{out}}$) denotes the
images whose perturbation is within the budget for DFC but not for the baseline (for
the baseline but not for DFC). Success rates (\%) of the baseline (B) and DFC (D) on
these images are measured regardless of the budget.}
  \label{tab:decomp}
  \setlength{\tabcolsep}{2pt}
  \renewcommand{\arraystretch}{1.1}
  \resizebox{\linewidth}{!}{
\begin{tabular}{lrrrcc}
\toprule
Method & Total & Density & Success & $C_{\mathrm{in}}$: B$\to$D & $C_{\mathrm{out}}$: B$\to$D \\
\midrule
GPG      & +5.78   & +2.60   & +3.19   & 20.22$\to$29.84 & 21.53$\to$27.28 \\
TSAA     & +9.31   & +3.79   & +5.52   & 18.24$\to$33.59 & 12.99$\to$19.30 \\
EGS-TSSA & $-$0.53 & $-$0.51 & $-$0.01 & 24.31$\to$17.79 & 13.13$\to$17.82 \\
\bottomrule
\end{tabular}
  }
  \vspace{-10pt}
\end{wraptable}

\paragraph{Where does the per-image advantage of DFC come from?}
DFC can gain under a per-image budget in two ways: by placing more images within
the budget, or by attacking the same images more successfully. We separate the
two by decomposing the difference in in-budget ASR between DFC and each baseline
into a density term, from images whose budget status changes, and a success term,
from changes in attack success on the same images; images whose budget status
changes also enter the success term with half weight (Table~\ref{tab:decomp};
App.~\ref{app:decomp}). For GPG and TSAA, both terms are positive, and the success term is the larger
(3.19 vs.\ 2.60~pp for GPG and 5.52 vs.\ 3.79~pp for TSAA).

The density term follows the shift in density: DFC lowers the density of both
attacks (Table~\ref{tab:baseline-dsc-cnn}, Fig.~\ref{fig:cdf}),
and the term is positive on all ten victims. This term thus reflects where each run ends relative to the budget, as
EGS-TSSA illustrates below, whereas the success term measures how effective
the perturbations are on the same images. The success term favors DFC whether it perturbs fewer or more pixels than the
baseline. On $C_{\mathrm{in}}$, where only DFC is within the budget and thus
perturbs fewer pixels, DFC succeeds more often (29.84\% vs.\ 20.22\% for GPG and
33.59\% vs.\ 18.24\% for TSAA); on $C_{\mathrm{out}}$, where it perturbs more, it
also succeeds more often (27.28\% vs.\ 21.53\% and 19.30\% vs.\ 12.99\%). On the images within the budget for both, which carry full weight in this term, DFC
succeeds 35.52\% vs.\ 26.64\% for GPG and 31.99\% vs.\ 18.76\% for TSAA. Unlike the
density term, the success term varies with the victim, consistent with the
transfer ASR without the budget (Tables~\ref{tab:baseline-dsc-cnn}
and~\ref{tab:baseline-dsc-transformer}): for GPG, it is negative on exactly the
four victims on which GPG-DFC has a lower transfer ASR than GPG (IncV3, VGG16,
Swin-T, and Twins-S).

EGS-TSSA shows the opposite pattern. Its DFC variant is denser (10.89\% vs.\
10.22\%), and the density term is negative on all ten victims. Its success also
follows density: on all ten victims, DFC succeeds less often than the baseline on
$C_{\mathrm{in}}$, where it is sparser, and more often on $C_{\mathrm{out}}$,
where it is denser. The success term is nearly zero ($-$0.01~pp), so the density
term accounts for almost all of the change in in-budget ASR ($-$0.51 of
$-$0.53~pp).

\subsection{\ours{} and the GFI Restriction}
\label{sec:gfi}

EGS-TSSA restricts the perturbations of TSAA to a GFI support derived from the
feature maps of the source model. Under a fixed coefficient, this restriction
improves TSAA: the mean transfer ASR rises from 25.23\% to 29.24\% on CNN
targets (Table~\ref{tab:baseline-dsc-cnn}), and the in-budget ASR at $\ell_0\le10\%$ from
6.41\% to 9.48\% (Table~\ref{tab:sparsity_transfer}). With \ours{}, it does not:
EGS-TSSA-\ours{} remains close to EGS-TSSA (29.54\% vs.\ 29.24\% on CNNs and
8.47\% vs.\ 8.87\% on Transformers) and falls well below TSAA-\ours{}
(40.83\% and 12.65\%).

\paragraph{What the restriction contributes.}
We decompose the change in in-budget ASR at $\ell_0\le10\%$ as in Table~\ref{tab:gfi-decomp}. Under a fixed
coefficient, the restriction raises the in-budget ASR of TSAA by 3.07~pp, of
which 2.14~pp comes from the density term: EGS-TSSA places 48.93\% of its
perturbations within the budget, compared with 37.40\% for TSAA. \ours{} alone
yields a larger density term (3.79~pp), together with a success term of
5.52~pp. With \ours{}, the restriction instead lowers the in-budget ASR by
6.76~pp, of which 5.20~pp comes from the success term: it mainly lowers attack
success on the same images.

\begin{wraptable}{r}{0.48\textwidth}
\vspace{-24pt}
\centering
\caption{Decomposition of the in-budget ASR change at $\ell_0 \le 10\%$,
averaged over the ten victims. Each row adds the component to the base;
$\rho$ is the mean density before $\rightarrow$ after (Table~\ref{tab:baseline-dsc-cnn}).}
\label{tab:gfi-decomp}
\footnotesize
\setlength{\tabcolsep}{2pt}
\begin{tabular}{lcrrr}
\toprule
 & & \multicolumn{3}{c}{ASR change (pp)} \\
\cmidrule(l){3-5}
Base & $\rho$ (\%) & Total & Dens. & Succ. \\
\midrule
\multicolumn{5}{l}{\textit{+ GFI mask}} \\
TSAA               & 10.86$\rightarrow$10.22        & $+3.07$ & $+2.14$ & $+0.94$ \\
TSAA-DFC & \phantom{0}9.99$\rightarrow$10.89 & $-6.76$ & $-1.56$ & $-5.20$ \\
\midrule
\multicolumn{5}{l}{\textit{+ DFC}} \\
TSAA               & 10.86$\rightarrow$\phantom{0}9.99 & $+9.31$ & $+3.79$ & $+5.52$ \\
\bottomrule
\end{tabular}
\vspace{-18pt}
\end{wraptable}

\paragraph{What density feedback cannot change.}
The GFI support $S(x)\in\{0,1\}^{H\times W}$ is computed from the clean image
and the fixed source model, and the restriction requires
$(1-S)\odot\delta=0$ regardless of the sparsity coefficient. \ours{} can
therefore change how many pixels the generator perturbs, but not which
locations it may use. Under a density target, the restriction thus acts in two
ways: it shifts the density, which \ours{} achieves on its own, and it
restricts the support, which \ours{} cannot relax. With \ours{}, the first
effect is redundant, while the second remains.

\paragraph{What the restriction removes.}
Evaluating the generator's restore perturbation (Sec.~\ref{sec:dynamics}) before the GFI mask
(App.~\ref{app:gfi}), we find that \ours{} makes the proposal of
EGS-TSSA more effective only outside the GFI support; within the support,
it is no more effective than under a fixed coefficient. The restriction
thus removes the gain that \ours{} brings. This is consistent with
Tab.~\ref{tab:gfi-decomp}, where the restriction lowers the in-budget ASR
of TSAA-\ours{} mainly through the success term.

\begin{figure}[t]
\centering
\includegraphics[width=\linewidth]{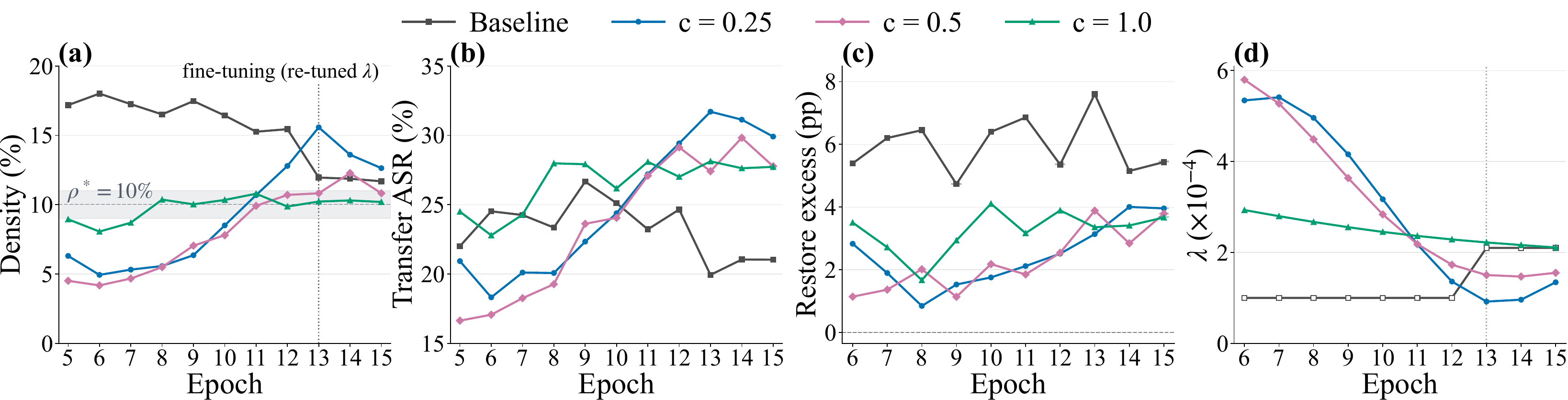}
\vspace{-18pt}
\caption{Training dynamics of GPG with a fixed coefficient (Baseline) and
GPG-\ours{} with damping $c$; the dotted line in (a) and (d) marks where the
baseline's coefficient is re-tuned for fine-tuning.
(a) Mean density, with $\rho^* = 10\% \pm 1$~pp shaded.
(b) Transfer ASR averaged over the ten victims.
(c) Restore excess: the ASR gain from restoring the locations discarded since
the previous epoch, minus that from adding as many random inactive locations
(App.~\ref{app:restore}).
(d) Sparsity coefficient $\lambda_t$.}
\label{fig:dynamics}
\vspace{-12pt}
\end{figure}

\subsection{Training Dynamics}
\label{sec:dynamics}

Fig.~\ref{fig:dynamics} traces GPG with a fixed coefficient and GPG-\ours{}
with damping $c\in\{0.25,0.5,1\}$ after warm-up. Transfer ASR is averaged over
the ten victims, and the final-epoch values of the baseline and of $c=0.5$
correspond to Tab.~\ref{tab:baseline-dsc-cnn} and~\ref{tab:baseline-dsc-transformer}.

\paragraph{Density.}
The baseline remains at 15--18\% during stabilization and approaches the
target only after its coefficient is re-tuned for fine-tuning at epoch 13
(Fig.~\ref{fig:dynamics}a). With $c=1$, \ours{} stays within one percentage
point of the target from epoch 8 onward. Smaller damping carries more
momentum: the density first undershoots to 4--5\% and then returns. With
$c=0.5$, it reaches the target band at epoch 11, briefly exceeds it at
epoch 14, and ends at 10.8\%; with $c=0.25$, it overshoots to 15.6\%
and ends at 12.6\%, outside the band. The damping coefficient thus shapes
the transient, and both $c=0.5$ and $c=1$ end within one percentage point of
the target with similar final ASR (27.76\% and 27.72\%).

\paragraph{Transferability recovers as the coefficient relaxes.}
The first updates of \ours{} tighten the coefficient enough to drive the
density below the target. Once $\lambda_t$ relaxes (Fig.~\ref{fig:dynamics}d),
the density recovers toward the target, and the ASR recovers with it
(Fig.~\ref{fig:dynamics}a,b): from epoch 6 until the density enters the
target band, it rises from 17.07\% to 27.08\% with $c=0.5$ and from 22.79\%
to 27.98\% with $c=1$. The recovered ASR overtakes the baseline while
\ours{} still perturbs fewer pixels: $c=1$ matches it at epoch 7 and exceeds
it from epoch 8, and $c\le0.5$ from epoch 11, when $c=0.5$ reaches 27.08\%
at a density of 9.91\% against 23.22\% at 15.26\% for the baseline. When
they overtake it, all three runs already exceed the highest ASR the baseline
reaches in Fig.~\ref{fig:dynamics}b (26.65\%), and they stay above it at
every later epoch except epoch 10 of $c = 1$. 
As the pressure relaxes, the ASR of \ours{} with $c=0.5$ reaches the level
of restoring what it discarded in the previous epoch, which the baseline
never reaches (Fig.~\ref{fig:current_restore}), suggesting that \ours{}
ends above the baseline by offsetting what it discards. The baseline instead approaches the target by raising its
coefficient, which lowers its ASR from 24.6\% to 19.9\% at fine-tuning.

\paragraph{What each epoch discards.}
The generators replace much of their support every epoch:
31--44\% of the active locations under a fixed coefficient and 35--66\%
with \ours{} (App.~\ref{app:restore}). To measure what they discard, we evaluate the perturbation values $P_t$ with the mask $M_t \lor M_{t-1}$, which restores the locations discarded since epoch $t-1$, and with $M_t \lor Q_t$, which instead activates as many random locations inactive in both epochs. The difference in ASR, the restore excess, thus reflects where the
discarded locations lie rather than how many there are
(Fig.~\ref{fig:dynamics}c). The baseline has a larger excess than all three
\ours{} runs at every epoch, even in epochs 6--9, when the density of \ours{}
recovers and it replaces up to 66\% of its support per epoch. The baseline's
excess averages 5.9~pp during stabilization, whereas that of DFC ($c \ge 0.5$) near the target (epochs 13--15) averages 3.5 pp despite replacing more.
Under a fixed coefficient, the generator thus discards more effective pixels.
Since random additions lower the ASR of every \ours{} run (App.~\ref{app:restore}), its rise during
density recovery likely comes from where the generator places the added
locations rather than from their number.

\section{Conclusion}
We proposed Density Feedback Control (\ours{}), which sets the sparsity coefficient of generator-based
sparse attacks by feedback from the hard-mask density instead of fixing it in
advance. In a single training run, \ours{} brings the mean density of three base
attacks to within one percentage point of the target, and for GPG and TSAA it
raises the mean transfer ASR on both CNN and Transformer targets at a lower
density, including under per-image $\ell_0$ budgets; with \ours{}, the earlier
TSAA surpasses the fixed-coefficient GPG. The training dynamics suggest why:
under a fixed coefficient, the density stays nearly constant after density
reduction and the ASR recovers little, whereas \ours{} relaxes the coefficient
once the density falls below the target, and the ASR recovers with the density,
overtaking the baseline with fewer pixels. For EGS-TSSA, the feature-importance
restriction helps under a fixed coefficient mainly by placing more images within
the budget, which \ours{} does on its own, while the restricted support removes
the gain of \ours{}. Our evaluation covers a single target density, source
model, and dataset, the training dynamics come from one seed of GPG, and
$\lambda_0$ was lowered for TSAA. Since \ours{} controls only the mean density
(Fig.~\ref{fig:cdf}), extending the feedback to per-image densities is left for
future work.

\newpage

\subsection*{AI use statement}

In this work, we used generative AI tools to assist with and polish the writing, support literature retrieval and discovery, and help draft and revise parts of the manuscript. We also used generative AI tools to assist with code generation and the presentation of experimental results in figures and tables to improve readability and visual clarity. We did not use generative AI tools for any other applicable tasks requiring disclosure.
We reviewed all AI-assisted work. In particular, we manually verified AI-assisted text and references, tested and checked generated code for correctness, and confirmed that all figures and tables accurately reflect the underlying experimental results. We take full responsibility for the final content of this work, including all text, claims, code, and artifacts produced with the assistance of generative AI.

\subsection*{Ethics statement}

This work investigates adversarial attack methods for evaluating model robustness. Although such techniques may have dual-use implications, the purpose of this research is to better understand model vulnerabilities and to facilitate the development and evaluation of more robust and secure learning systems.

\subsection*{Reproducibility statement}

We provide the algorithmic details of our method in the main paper and report the datasets, model configurations, training settings, and evaluation protocols used in our experiments. Additional implementation details and hyperparameters required to reproduce our results are provided. Code will be made publicly available.



\bibliography{iclr2027_conference}
\bibliographystyle{iclr2027_conference}

\newpage
\appendix
\section{Restore Experiment}
\label{app:restore}

This appendix defines the restore experiment of Sec.~\ref{sec:dynamics} and
reports its per-epoch results (Table~\ref{tab:restore}).

\paragraph{Masks and replacement.}
For a run and epoch $t$, let $M_t(x)\in\{0,1\}^{H\times W}$ denote the hard
spatial mask produced by the generator at the end of epoch $t$, where a
location is active if any color channel is perturbed, and let $P_t(x)$ denote
its perturbation values before masking, so that the current perturbation is
$\delta_t=M_t\odot P_t$. The locations discarded and added between epochs
$t-1$ and $t$ are
\begin{equation}
R_t=M_{t-1}\land\lnot M_t,\qquad A_t=M_t\land\lnot M_{t-1}.
\end{equation}
We report $\rho_t$, $r_t$, and $a_t$ as the fractions of the $HW$ spatial
locations that are in $M_t$, $R_t$, and $A_t$, averaged over images. The
replacement rate $r_t/\rho_{t-1}$ is the fraction of the locations active at
epoch $t-1$ that are inactive at epoch $t$, computed from these averages with
$\rho_{t-1}=\rho_t+r_t-a_t$.

\paragraph{Restore and random conditions.}
Both conditions evaluate the current perturbation values on an enlarged mask:
\begin{align}
\label{eq:restore}
x_{\mathrm{restore}} &= \mathrm{clip}\bigl(x+P_t\odot(M_t\lor M_{t-1}),\,0,\,1\bigr),\\
x_{\mathrm{random}}  &= \mathrm{clip}\bigl(x+P_t\odot(M_t\lor Q_t),\,0,\,1\bigr),
\end{align}
where, for each image, $Q_t$ is drawn uniformly without replacement from the
locations inactive in both epochs, $\lnot(M_t\lor M_{t-1})$, with
$|Q_t|=|R_t|$. Both conditions therefore perturb $|M_t|+|R_t|$ locations on
each image, and their mean density is $\rho_t+r_t$. We define
\begin{equation}
G_{\mathrm{restore}}=\mathrm{ASR}_{\mathrm{restore}}-\mathrm{ASR}_{\mathrm{cur}},\qquad
G_{\mathrm{random}}=\mathrm{ASR}_{\mathrm{random}}-\mathrm{ASR}_{\mathrm{cur}},
\end{equation}
and the restore excess as $G_{\mathrm{restore}}-G_{\mathrm{random}}
=\mathrm{ASR}_{\mathrm{restore}}-\mathrm{ASR}_{\mathrm{random}}$.

\paragraph{Protocol.}
We use the checkpoints of adjacent epochs $t-1$ and $t$ for $t=6,\dots,15$. Epoch 5 is omitted because, for the baseline, its previous epoch belongs to
density reduction; we use the same epochs for all runs. All conditions are evaluated on the 50{,}000 ImageNet validation
images with $\epsilon=10/255$; for each victim, ASR is measured on the images
it classifies correctly, and the ten victims are weighted equally. $Q_t$ is
drawn with a single random seed shared by all runs, so we do not report
variation across seeds. At epoch 11 of the baseline, one image has fewer
candidate locations than $|R_t|$; we exclude it from all three conditions of
that row (49{,}999 images). Densities are computed over all 50{,}000 images.
The epoch-15 values of $\mathrm{ASR}_{\mathrm{cur}}$ for the baseline and for
$c=0.5$ match Tables~\ref{tab:baseline-dsc-cnn} and~\ref{tab:baseline-dsc-transformer}.

\begin{figure}[t]
\centering
\includegraphics[width=\linewidth]{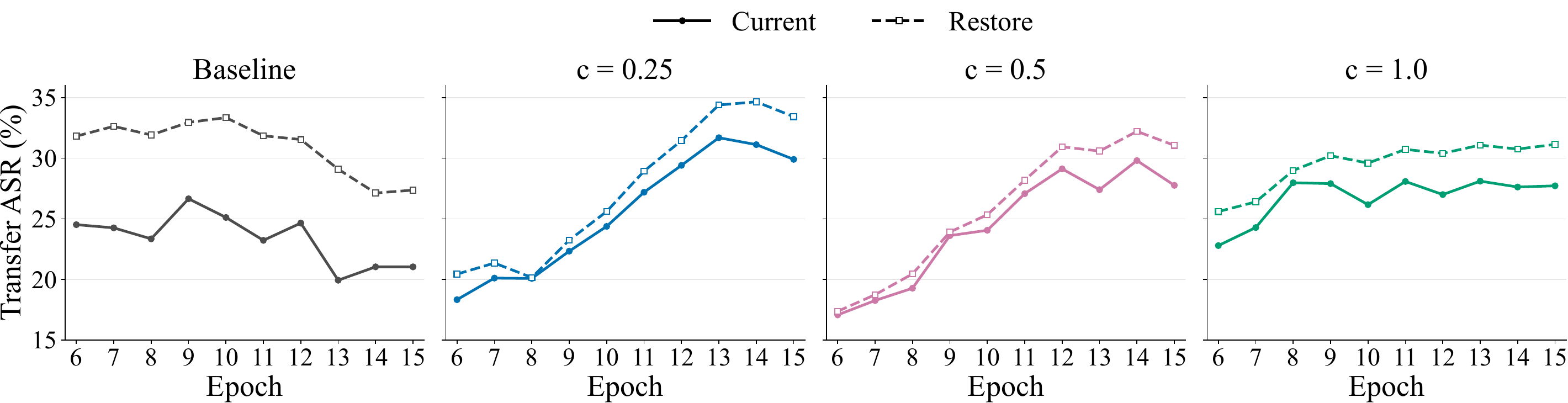}
\caption{Current ASR (solid) and restore ASR (dashed) of GPG with a fixed
coefficient (Baseline) and GPG-\ours{} with damping $c$, averaged over the
ten victims. The restore ASR evaluates the current perturbation after
restoring the locations discarded in the previous epoch
(Eq.~\ref{eq:restore}). With $c=0.5$, the current ASR at epochs 7--12
reaches the restore ASR of the previous epoch, whereas the baseline never
reaches its restore level.}
\label{fig:current_restore}
\end{figure}

\paragraph{Results.}
The replacement rate ranges from 31 to 44\% for the baseline and from 35 to
66\% with \ours{}; the highest rates occur while the density of $c\le0.5$
recovers from the undershoot (epochs 6--9). Random additions raise the ASR of
the baseline by 0.90--2.18~pp and lower that of every \ours{} run by
0.25--0.89~pp. The excess of the baseline exceeds that of all three \ours{}
runs at every epoch, by at least 1.15~pp (epoch 14, $c=0.25$), and its
smallest value (4.73~pp, epoch 9) is larger than the largest value of any
\ours{} run (4.11~pp, $c=1$ at epoch 10). Fig.~\ref{fig:current_restore} plots the current and restore ASR of each run.

\begin{table}[h]
\centering
\caption{Per-epoch restore results for GPG with a fixed coefficient (Baseline)
and GPG-\ours{} with damping $c$. Densities, replacement rates, and ASR are in
\%; gains and excess are in pp. Both the restore and random conditions have a
mean density of $\rho_t+r_t$. ASR is averaged over the ten victims.}
\label{tab:restore}
\small
\setlength{\tabcolsep}{4pt}
\begin{tabular}{llrrrrrrrr}
\toprule
Run & $t$ & $\rho_t$ & $r_t$ & $a_t$ & $r_t/\rho_{t-1}$ &
$\mathrm{ASR}_{\mathrm{cur}}$ & $G_{\mathrm{restore}}$ &
$G_{\mathrm{random}}$ & Excess \\
\midrule
Baseline & 6 & 18.01 & 5.89 & 6.72 & 34.29 & 24.52 & 7.31 & 1.91 & 5.39 \\
 & 7 & 17.25 & 6.67 & 5.91 & 37.06 & 24.25 & 8.38 & 2.18 & 6.20 \\
 & 8 & 16.51 & 6.74 & 6.00 & 39.09 & 23.35 & 8.58 & 2.12 & 6.45 \\
 & 9 & 17.47 & 5.11 & 6.07 & 30.96 & 26.65 & 6.31 & 1.57 & 4.73 \\
 & 10 & 16.43 & 6.41 & 5.36 & 36.67 & 25.11 & 8.24 & 1.84 & 6.40 \\
 & 11 & 15.26 & 6.10 & 4.94 & 37.14 & 23.22 & 8.63 & 1.77 & 6.86 \\
 & 12 & 15.44 & 5.10 & 5.28 & 33.43 & 24.65 & 6.90 & 1.55 & 5.35 \\
 & 13 & 11.95 & 6.74 & 3.25 & 43.63 & 19.93 & 9.17 & 1.57 & 7.60 \\
 & 14 & 11.87 & 4.32 & 4.24 & 36.15 & 21.04 & 6.09 & 0.94 & 5.15 \\
 & 15 & 11.69 & 4.37 & 4.18 & 36.77 & 21.03 & 6.33 & 0.90 & 5.43 \\
\midrule
$c=0.25$ & 6 & 4.94 & 4.07 & 2.70 & 64.48 & 18.33 & 2.11 & $-$0.72 & 2.83 \\
 & 7 & 5.32 & 2.68 & 3.06 & 54.27 & 20.11 & 1.24 & $-$0.65 & 1.89 \\
 & 8 & 5.58 & 3.52 & 3.78 & 66.14 & 20.08 & 0.07 & $-$0.77 & 0.84 \\
 & 9 & 6.36 & 3.13 & 3.91 & 56.21 & 22.33 & 0.91 & $-$0.62 & 1.53 \\
 & 10 & 8.51 & 2.81 & 4.96 & 44.25 & 24.38 & 1.23 & $-$0.53 & 1.76 \\
 & 11 & 10.69 & 3.31 & 5.50 & 38.95 & 27.19 & 1.74 & $-$0.38 & 2.12 \\
 & 12 & 12.79 & 4.08 & 6.18 & 38.13 & 29.41 & 2.05 & $-$0.46 & 2.51 \\
 & 13 & 15.58 & 4.44 & 7.23 & 34.73 & 31.70 & 2.70 & $-$0.44 & 3.14 \\
 & 14 & 13.60 & 6.62 & 4.65 & 42.53 & 31.12 & 3.53 & $-$0.47 & 4.00 \\
 & 15 & 12.63 & 5.70 & 4.73 & 41.92 & 29.91 & 3.53 & $-$0.43 & 3.96 \\
\midrule
$c=0.5$ & 6 & 4.18 & 2.66 & 2.33 & 58.92 & 17.07 & 0.30 & $-$0.84 & 1.14 \\
 & 7 & 4.68 & 2.32 & 2.81 & 55.39 & 18.25 & 0.48 & $-$0.88 & 1.36 \\
 & 8 & 5.50 & 2.39 & 3.21 & 51.14 & 19.27 & 1.18 & $-$0.83 & 2.02 \\
 & 9 & 7.05 & 2.90 & 4.45 & 52.71 & 23.61 & 0.30 & $-$0.84 & 1.14 \\
 & 10 & 7.80 & 3.46 & 4.21 & 49.11 & 24.05 & 1.28 & $-$0.89 & 2.18 \\
 & 11 & 9.91 & 3.52 & 5.64 & 45.19 & 27.08 & 1.11 & $-$0.74 & 1.85 \\
 & 12 & 10.70 & 4.21 & 5.00 & 42.45 & 29.12 & 1.82 & $-$0.72 & 2.54 \\
 & 13 & 10.82 & 4.90 & 5.01 & 45.75 & 27.40 & 3.19 & $-$0.70 & 3.88 \\
 & 14 & 12.27 & 3.77 & 5.22 & 34.83 & 29.81 & 2.41 & $-$0.43 & 2.84 \\
 & 15 & 10.82 & 6.09 & 4.64 & 49.59 & 27.76 & 3.30 & $-$0.49 & 3.79 \\
\midrule
$c=1$ & 6 & 8.06 & 4.69 & 3.80 & 52.41 & 22.79 & 2.80 & $-$0.70 & 3.50 \\
 & 7 & 8.70 & 3.48 & 4.11 & 43.12 & 24.28 & 2.12 & $-$0.59 & 2.72 \\
 & 8 & 10.37 & 3.81 & 5.47 & 43.76 & 27.98 & 1.01 & $-$0.65 & 1.67 \\
 & 9 & 10.03 & 4.53 & 4.19 & 43.66 & 27.91 & 2.31 & $-$0.63 & 2.94 \\
 & 10 & 10.34 & 4.53 & 4.84 & 45.17 & 26.17 & 3.43 & $-$0.67 & 4.11 \\
 & 11 & 10.79 & 4.68 & 5.13 & 45.24 & 28.08 & 2.65 & $-$0.52 & 3.16 \\
 & 12 & 9.87 & 5.11 & 4.19 & 47.37 & 27.00 & 3.40 & $-$0.49 & 3.89 \\
 & 13 & 10.22 & 3.94 & 4.30 & 39.92 & 28.11 & 2.98 & $-$0.38 & 3.36 \\
 & 14 & 10.31 & 4.05 & 4.13 & 39.60 & 27.63 & 3.13 & $-$0.28 & 3.41 \\
 & 15 & 10.19 & 4.32 & 4.21 & 41.95 & 27.72 & 3.42 & $-$0.25 & 3.67 \\
\bottomrule
\end{tabular}
\end{table}

\section{GFI Restriction on the Restore Perturbation}
\label{app:gfi}

This appendix details the evaluation in Sec.~\ref{sec:gfi} and reports
its results (Table~\ref{tab:gfi_restore}).

\paragraph{Proposal and restriction.}
The generator of EGS-TSSA outputs a mask $M_t$ and perturbation values $P_t$
over the whole image, and the GFI support $S(x)\in\{0,1\}^{H\times W}$,
computed from the clean image and the fixed source model, is applied
afterwards, so the perturbation it uses is $S\odot M_t\odot P_t$. We call
$M_t\odot P_t$, before $S$ is applied, the proposal of the generator.

\paragraph{Conditions.}
Since the current ASR of GPG-\ours{} follows the restore ASR
(Sec.~\ref{sec:dynamics}), we ask how much of the restore ASR depends on
locations outside the support. We evaluate the restore perturbation of
Eq.~\ref{eq:restore}, with $M_t$ and $M_{t-1}$ taken before $S$, without
and with $S$:
\begin{equation}
  x_{\mathrm{restore}}^{S}
  = \mathrm{clip}\bigl(x + S\odot P_t\odot(M_t\lor M_{t-1}),\,0,\,1\bigr).
\end{equation}
The drop
$\mathrm{ASR}_{\mathrm{restore}}-\mathrm{ASR}_{\mathrm{restore}}^{S}$
measures how much of the restore ASR depends on locations outside the
support.

\paragraph{Protocol.}
We use the checkpoints of epochs 7--14, over which \ours{} relaxes the
sparsity pressure and the transfer ASR of EGS-TSSA-\ours{} recovers, for
both EGS-TSSA and EGS-TSSA-\ours{}. All other settings follow
App.~\ref{app:restore}: the 50,000 ImageNet validation images,
$\epsilon=10/255$, ASR measured on the images each victim classifies
correctly, and the ten victims weighted equally. We average the ASR over
these epochs of a single training run.

\paragraph{Results.}
Without $S$, the proposal of EGS-TSSA-\ours{} is more effective than that
of EGS-TSSA (25.36\% vs.\ 22.44\%); with $S$, it is not (16.74\% vs.\
17.19\%). The mask thus lowers the restore ASR by 8.62~pp for
EGS-TSSA-\ours{}, against 5.25~pp for EGS-TSSA. The gain that \ours{}
brings to the proposal lies outside the support, and the restriction
removes it.

\begin{table}[h]
\centering
\caption{Restore ASR (\%) of EGS-TSSA and EGS-TSSA-\ours{} without and with
the GFI support $S$, averaged over the ten victims and epochs 7--14.
Without $S$, the restore perturbation is evaluated on the proposal before
the GFI mask.}
\label{tab:gfi_restore}
\begin{tabular}{lrrr}
\toprule
Method & Without $S$ & With $S$ & Drop (pp) \\
\midrule
EGS-TSSA         & 22.44 & 17.19 & 5.25 \\
EGS-TSSA-\ours{} & 25.36 & 16.74 & 8.62 \\
\bottomrule
\end{tabular}
\end{table}

\section{Decomposition of the Per-Image Gain}
\label{app:decomp}

For a victim model and a correctly classified image $i$, let
$b_i^M = \indic[\rho_i^M \le \tau]$ indicate whether the perturbation of
method $M$ is within the $\ell_{0}$ budget $\tau$, and let $s_i^M$ indicate whether it
fools the victim regardless of the budget. The in-budget ASR is
$\mathrm{ASR}^M = \frac{1}{N}\sum_i b_i^M s_i^M$. For DFC ($D$) and a baseline
($B$),
\begin{equation}
\mathrm{ASR}^D - \mathrm{ASR}^B
= \underbrace{\frac{1}{N}\sum_i \bigl(b_i^D - b_i^B\bigr)\frac{s_i^D + s_i^B}{2}}_{\text{density term}}
+ \underbrace{\frac{1}{N}\sum_i \bigl(s_i^D - s_i^B\bigr)\frac{b_i^D + b_i^B}{2}}_{\text{success term}}.
\end{equation}
The density term is nonzero only for images whose budget status differs between
the two methods, and the success term only for images on which attack success
differs. The decomposition is exact and symmetric in the two methods. We compute
it for each victim model and report the average over the ten victims.

\newpage
\section{Algorithm}\label{app:alg}
Algorithm~\ref{alg:dfc} summarizes DFC. The controller runs outside the
generator's computational graph and adds no trainable parameters.

\begin{algorithm}[h]
\caption{Density Feedback Control (\ours{})}
\label{alg:dfc}
\begin{algorithmic}[1]
\Require target density $\rho^*$, initial coefficient $\lambda_0$, damping $c$,
         warm-up epochs $T_w$, total epochs $T$
\State $\kappa \gets 10^{\lfloor\log_{10}\lambda_0\rfloor}$;\ $v \gets 0$;\ $\lambda \gets \lambda_0$
\State Train $G_\theta$ for $T_w$ epochs with $\lambda=0$ \Comment{warm-up}
\For{$t = T_w+1, \dots, T$}
  \State Train $G_\theta$ for one epoch on $\mathcal{L}_{\mathrm{adv}}+\lambda\mathcal{L}_{\mathrm{spa}}$,
         recording hard-mask densities
  \State $\rho_t \gets$ mean density over the epoch;\ $e_t \gets (\rho_t-\rho^*)/\rho^*$
\State $v \gets (1-c)\,v + \kappa e_t$; $\lambda \gets \max(0,\ \lambda + v)$ \Comment{keep $\lambda \ge 0$}
\EndFor
\end{algorithmic}
\end{algorithm}

\section{Training details of generator-based attacks}\label{app:gen}
All generator-based methods are trained for 15 epochs and start with a warm-up
with $\lambda = 0$. For the baselines, a three-epoch warm-up is followed by
reduction and stabilization phases spanning nine epochs and three epochs of
fine-tuning toward a target density of 10\%, each with a fixed $\lambda$
(Table~\ref{tab:gen-coef}). For the DFC variants, we shorten the warm-up to two
epochs so that the density can settle at the target within the same 15-epoch
budget, and DFC replaces the remaining phases. We initialize $v_0 = 0$ and
update the sparsity coefficient at the end of each epoch toward
$\rho^* = 10\%$. For GPG and EGS-TSSA, we set $\lambda_0$ to $10^{-4}$, the
default sparsity coefficient of GPG. For TSAA, this setting drove the density
to zero after warm-up, discarding what the generator had learned during
warm-up; we therefore lower $\lambda_0$ to $5 \times 10^{-5}$.

\begin{table}[h]
\centering
\caption{Sparsity coefficients of the generator-based attacks.}
\label{tab:gen-coef}
\begin{tabular}{lcccc}
\toprule
 & \multicolumn{2}{c}{Baseline $\lambda$} & \multicolumn{2}{c}{\ours{}} \\
\cmidrule(lr){2-3}\cmidrule(lr){4-5}
Method & Red./Stab. & Fine-tune & $\lambda_0$ & $\kappa$ \\
\midrule
GPG      & $1\times10^{-4}$   & $2.1\times10^{-4}$ & $10^{-4}$        & $10^{-4}$ \\
TSAA     & $1.5\times10^{-4}$ & $2\times10^{-4}$   & $5\times10^{-5}$ & $10^{-5}$ \\
EGS-TSSA & $8\times10^{-5}$   & $2\times10^{-5}$   & $10^{-4}$        & $10^{-4}$ \\
\bottomrule
\end{tabular}
\end{table}

\section{Limitations}
\label{app:limitations}

\paragraph{Strict per-image budgets.}
Because \ours{} controls only the mean density, the density still varies across images, and perturbations
whose density exceeds a per-image $\ell_0$ budget do not satisfy the constraint.
\ours{} improves transferability under per-image budgets over the
fixed-coefficient baselines (Table~\ref{tab:sparsity_transfer}), but under strict
budgets such as $\ell_0 \leq 10\%$, its attack success rate remains below that
obtained without the per-image constraint at the same mean density. Closing this
gap, for example by feeding back an upper quantile of the per-image density
distribution instead of the mean, is left for future work.

\paragraph{Restricted perturbation regions.}
\ours{} improves transferability by letting the generator re-select pixels when
the sparsity pressure is relaxed. When the base attack restricts perturbations to
a predefined region, as EGS-TSSA does with feature-importance regions of the
source model, this re-selection is confined to the region, and \ours{} yields no
consistent gain (Tables~\ref{tab:baseline-dsc-cnn} and~\ref{tab:baseline-dsc-transformer}). As
analyzed in Section~\ref{sec:gfi}, a region that benefits fixed-coefficient
training becomes a constraint once pixel selection is driven by density feedback.
Combining density feedback with such region restrictions is left for future work.

\end{document}